# DOUBLE FOUR-BAR CRANK-SLIDER MECHANISM DYNAMIC BALANCING BY META-HEURISTIC ALGORITHMS

Habib Emdadi[1], Mahsa Yazdanian[2], Mir Mohammad Ettefagh[3] and Mohammad-Reza Feizi-Derakhshi[4]

[1,4]Department of Computer Engineering, University of Tabriz, Tabriz, Iran
[2,3]Department of Mechanical Engineering, University of Tabriz, Tabriz, Iran

## ABSTRACT

*In this paper, a new method for dynamic balancing of double four-bar crank slider mechanism by meta-heuristic-based optimization algorithms is proposed. For this purpose, a proper objective function which is necessary for balancing of this mechanism and corresponding constraints has been obtained by dynamic modeling of the mechanism. Then PSO, ABC, BGA and HGAPSO algorithms have been applied for minimizing the defined cost function in optimization step. The optimization results have been studied completely by extracting the cost function, fitness, convergence speed and runtime values of applied algorithms. It has been shown that PSO and ABC are more efficient than BGA and HGAPSO in terms of convergence speed and result quality. Also, a laboratory scale experimental doublefour-bar crank-slider mechanism was provided for validating the proposed balancing method practically.*

## 1. INTRODUCTION

Dynamic balancing is one of the important concerns in high speed mechanisms such as crank slider mechanism, a special attention should paid to the inertia-induced force (shaking force) and moment (shaking moment) transmitted to the frame in order to reduce vibration amplitude. The methods of balancing linkages are well developed and documented in [1]. These techniques mostly are based on mass redistribution, addition of counterweights to the moving links, and attachment of rotating disks or duplication of the linkages [2]. In these methods, the shaking forces and shaking moments should be minimized. One of these methods is "Maximum recursive dynamic algorithm" that published by Chaudhary and Saha [3]. Another method which is documented by QI and Pennestrlis [4] is called "refined algorithm". It presents a numerically efficient technique for the optimum balancing of linkages. In this approach, instead of solving directly the dynamic equations, a technique is introduced to solve the linked dynamic equations in a "shoe string" fashion. The comprehensive mass distribution method, for an optimum balancing of the shaking force and shaking moment is used by Yu [5] to optimal balancing of the spatial RSSR mechanism.

In recent years, optimization techniques have received widespread interest in the engineering sciences. These techniques were initiated, primarily, by the advent of high speed digital computers which give a practical, economical and accurate means of obtaining solutions to meaningful problems in the optimization of static and dynamic systems. Dynamic balancing inherently constitutes an optimization problem. For instance, CWB (counterweight balancing) involves a trade-off between minimizing the different dynamic re-actions. Therefore, determining the counterweights' mass parameters is an optimization problem. Also, Alici and Shirinzadeh [6] considered "Sensitivity analysis". In this technique they formulated the dynamic balancing as an optimization problem such that while the shaking force balancing is accomplished through analytically obtained balancing constraints, an objective function based on the sensitivity analysis of shaking moment with respect to the position, velocity and acceleration of the links is used to minimize the shaking moment. Evolutionary algorithms such as GA and PSO are useful in optimization problems, because they are easy to implement for the complex real-world problems. In some cases that objective function could not expressed as an explicit function of design variables, using these evolutionary algorithms is more suitable for optimization than the traditional deterministic optimization methods. Wen-Yi Lin proposed a GA-DE hybrid algorithm for application to path synthesis of a four-bar linkage [7]. Acharyya and Mandal[8]applied BGA with multipoint crossover, the PSO with the CFA (constriction factor approach) and the DE to the path synthesis of a four-bar linkage. Jaamiolahmadi and Farmani [9] proposed another related work. They described the application of GA to force and moment-balance of a four bar linkage.

The aim of present work is applying new artificial intelligent based optimization methods for balancing double four-bar crank slider mechanism which is a special case of four-bar mechanism, but it has its own specification in dynamic balancing. Ettefagh and Abbasidoust [10] applied BGA for balancing of a crank slider mechanism, however in present work a comprehensive study of different algorithms proposed for balancing of a crank slider mechanism was carried out. The major objective of this paper is to compare the computational effectiveness and efficiency of BGA (Binary Genetic Algorithm), PSO (Particle Swarm Optimization), ABC (Artificial Bee Colony) and HGAPSO (Hybrid GA and PSO) algorithms. The performance of the optimization techniques in terms of computational time and convergence rate is compared in the results section. After kinematics formulating of the mechanism and extracting the object function with its' constraints, results of the proposed method on a simulated mechanism will be reported. Finally, for experimentally validation of the method, a laboratory scale experimental crank slider mechanism will be considered and the proposed optimization method will be applied for its balancing.

## 2. MODELLING

The selected mechanism in this paper is shown in Figure 1, which is a benchmark double four-bar mechanism and this mechanism is applied in different machines [10]. Therefore the proposed balancing method in this paper may be easily applied to other mechanisms in different machines for balancing. As shown in Figure 1, the mechanism contains two cranks-sliders and four rotating disks. Each rode is attached to disk 1 and disk 4. All disks are connected to a shaft which is rotating with an electromotor. When the mechanism is working, shaking force and moments cause frame to vibrate. Our goal is to minimize shaking forces and moments according to the regarded constraints. The mechanism is modeled as shown in Figure 2 by computing shaking forces of the mechanism on all four planes.

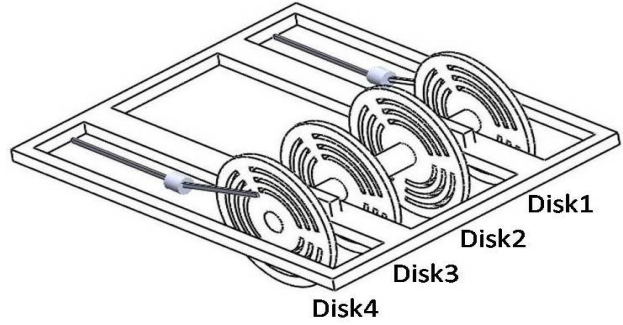

Figure 1.Schematic diagram of the considered double four-bar crank-slider mechanism [10]

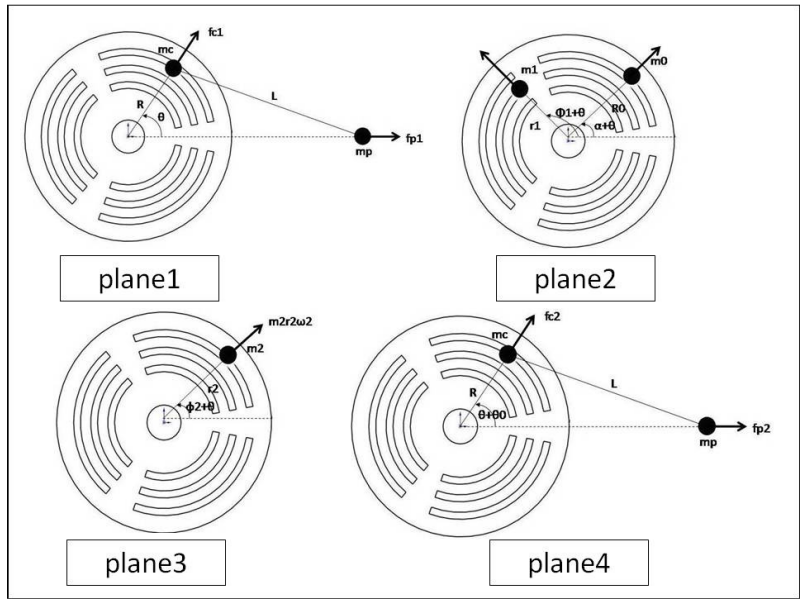

Figure 2.Shaking forces of the mechanism on all four plans [10]

In this paper, the method used for balancing is the addition of two counterweights on disks 2 and 3. In addition, system has an unbalance mass on disk 2. Shaking forces in crank slider as the dynamical forces of mechanism are formulated as:

$$f_c = m_c R \omega^2 \qquad (1)$$

$$f_p = m_p R \omega^2 (\cos\theta + \frac{R}{L}\cos 2\theta) \qquad (2)$$

Where $m_c$ is the eccentric mass and $m_p$ is the equal mass of the slider, $R$ is corresponding radius and $\omega$ is mechanism rotating speed. Also, $\theta$ indicates the angular position of the disk and $L$ is the connecting rod's length. Equilibrium on $x$ and $y$ axes due to the acting forces are described by:

$$\sum F_x = 0 \rightarrow$$
$$m_p R\omega^2 (\cos\theta + \frac{R}{L}\cos 2\theta) + m_c R\omega^2 \cos\theta + m_0 R_0 \omega^2 \cos(\theta + \alpha) + m_1 r_1 \omega^2 \cos(\theta + \varphi_1) + m_2 r_2 \omega^2 \cos(\theta + \varphi_2) +$$
$$m_p R\omega^2 (\cos(\theta + \theta_0) + \frac{R}{L}\cos 2(\theta + \theta_0)) + m_c R\omega^2 \cos(\theta + \theta_0) = 0 \qquad (3)$$

$$\sum F_y = 0 \rightarrow$$
$$m_c R\omega^2 \sin\theta + m_0 R_0 \omega^2 \sin(\theta + \alpha) + m_1 r_1 \omega^2 \sin(\theta + \varphi_1) + m_2 r_2 \omega^2 \sin(\theta + \varphi_2) +$$
$$m_c R\omega^2 \sin(\theta + \theta_0) = 0 \qquad (4)$$

Also Momentum equilibrium equations are written as:

$$\sum M_y = 0 \rightarrow$$
$$[m_0 R_0 \omega^2 \cos(\theta + \alpha) + m_1 r_1 \omega^2 \cos(\theta + \varphi_1)]a_1 + [m_2 r_2 \omega^2 \cos(\theta + \varphi_2)](a_1 + a_2) +$$
$$\left[ m_p R\omega^2 (\cos(\theta + \theta_0) + \frac{R}{L}\cos 2(\theta + \theta_0)) \right](2a_1 + a_2) + [m_c R\omega^2 \cos(\theta + \theta_0)](2a_1 + a_2) = 0 \qquad (5)$$

$$\sum M_x = 0 \rightarrow$$
$$[m_0 R_0 \omega^2 \sin(\theta + \alpha) + m_1 r_1 \omega^2 \sin(\theta + \varphi_1)]a_1 + [m_2 r_2 \omega^2 \sin(\theta + \varphi_2)](a_1 + a_2) +$$
$$[m_c R\omega^2 \sin(\theta + \theta_0)](2a_1 + a_2) = 0 \qquad (6)$$

$m_0$ is the constrain mass on disk 2, $R_0$ and $\alpha$ are corresponding radius and angular position of the mass, respectively.

In construction of optimization problems, as described before, some parameters of system is applies as the variables of cost function to minimize it. In our method, $m_1$, $m_2$, $\varphi_1$, $\varphi_2$, $r_1$ and $r_2$ are counterweights parameters that are unknown. We take $r_1$ and $r_2$ as known parameters. So we define the cost function as below. The cost function is consisting of forces equations and the constraints are considered by moment equations. For this purpose, we define some parameters as $P_1$, $P_2$, $P_3$ and $P_4$ as below:

$$P_1(\theta) = \sum F_x \quad , \quad P_2(\theta) = \sum F_y \qquad (7)$$

$$P_3(\theta) = \sum M_x \quad , \quad P_4(\theta) = \sum M_y \qquad (8)$$

These parameters are the sum of forces and moments in $x$, $y$ direction. Consequently, the cost function is:

$$\text{Cost function: } f(m_1, m_2, \varphi_1, \varphi_2) = \int \{|P_1(\theta)| + |P_2(\theta)|\} dA \qquad (9)$$

The cost function is a function of the parameters $m_1$, $m_2$, $\varphi_1$ and $\varphi_2$, which should be found. In other words, the cost function is the area that $\{|P_1(\theta)| + |P_2(\theta)|\}$ makes in polar system. Our goal is to minimize this area. Additionally, by defining C1 and C2 as follows:

$$C1(m_1, m_2, \varphi_1, \varphi_2) = \int |P_3(\theta)| dA \qquad (10)$$

$$C2(m_1, m_2, \varphi_1, \varphi_2) = \int |P_4(\theta)| dA \qquad (11)$$

Constraints are considered such that C1 and C2 to be less than predefined values which determined by searching among the maximum value of them during running the Meta-Heuristic Algorithms.

## 3. APPLYING META-HEURISTIC ALGORITHMS

PSO, ABC and HGAPSO algorithms, which is introduced in summary in following section, are applied for minimizing the mentioned cost function with defined constraints as it was described in previous section.

### 3.1. Particle Swarm Optimization algorithm

The PSO method introduced for the first time in 1995 by Kennedy and Eberhart [11]. This method based on social behavior of organisms such as fish schooling and bird flocking. Because of its simple concept and quick convergence, PSO can be applied to various applications in different fields. The approach uses the concept of population and a measure of performance like the fitness value used with other evolutionary algorithms. Similar to other evolutionary algorithms, the system is initialized with a population of random solutions, called particles.
Each particle maintains its own current position, its present velocity and its personal best position explored so far. The swarm is also aware of the global best position achieved by all its members. The iterative appliance of update rules leads to stochastic manipulation of velocities and flying courses. During the process of optimization the particles explore the *D*-dimensional space, whereas their trajectories can probably depend both on their personal experiences, on those of their neighbors and the whole swarm, respectively. In every iteration, each particle is updated by following the best solution of current particle achieved so far (*Pbest*) and the best of the population (*gbest*). When a particle takes part of the population as its topological neighbours, the best value is a local best. The particles tend to move to good areas in the search space by the information spreading to the swarm. The particle is moved to a new position calculated by the velocity updated at each time step t. This new position is calculated as the sum of the previous position and the new velocity by equation (12).

$$v_i^{t+1} = wv_i^t + c_i rand(0,1)(Pbest_i - x_i^t) + c_i rand(0,1)(gbest_t - x_i^t) \qquad (12)$$

The velocity updated as following equation:
$$x_i^{t+1} = x_i^t + v_i^{t+1} \qquad (13)$$

The parameter *w* is called the inertia weight and controls the magnitude of the old velocity $v_i^t$ in the calculation of the new velocity which is $v_i^{t+1}$, whereas the acceleration factors $C_1$ and $C_2$ determine the significance of *Pbest* and *gbest* respectively. In the present work we selected $C_1$=0.25 and $C_2$ =0.15. Furthermore, $x_i^t$ is the current particle position and $x_i^{t+1}$ is the new position of the particle. $rand(0,1)$ is the random number uniformly distributed. In general, the inertia weight *w* is set according to the following equation:

$$w = w_{max} - \left(\frac{w_{max} - w_{min}}{Ir_{max}}\right) \times Iter \qquad (14)$$

Where $w_{max}$ is initial weight, $w_{min}$ is final weight, $Ir_{max}$ is the maximum number of iteration (generation) and $Iter$ is current iteration number.

Main steps of the procedure are:
1. Position and initial velocity of particles are generated randomly.
2. The objective function value i calculated for each particle.

3. $Pbest$ For each particle is considered equal to its initial position (In the first iteration). Also $gbest$ determined.
4. Modify position and velocity of each particle using the equations (12) and (13).
5. Calculate the overall objective function for each particle.
6. If the value of new overall objective function is better than the value of each particle in *P-best*, then it would be replaced in *P-best* and if in whole new population gets found which is better fitting in objective function would be replaced by *g-best*.
7. Repeat steps 4 to 7 to reach the maximum number specified. The final answer is obtained from the $gbest$ in the last iteration.

Figure 3 shows the flowchart of the developed PSO algorithm [11].

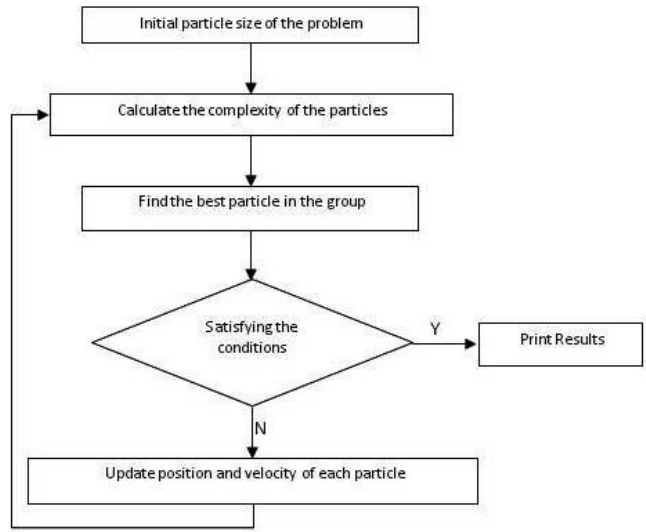

Figure 3. Flowchart of the PSO algorithm

## 3.2. Artificial Bee Colony algorithm

The artificial bee colony (ABC) algorithm is a new population-based stochastic heuristic approach which was introduced for the first time by Karaboga in 2005 [12]. This algorithm is very simple and flexible, which does not require external parameters like crossover rates, especially suitable for engineering application. The colony of artificial bees consists of three groups of bees to search foods generally, which includes employed bees, onlookers and scouts. The first half of colony consists of the employed artificial bees and the second half includes the onlookers. The employed bees search the food around the food source in their memory. They perform waggle dance upon returning to the hive to pass their food information to the other of the colony (the onlookers). The onlookers are waiting around the dance floor to choose any of the employed bees to follow based on the nectar amount information shared by the employed bees. The employed bee whose food source has been exhausted by the bees becomes a scout who must find new food source. For every food source, there is only one employed bee. In other words, the number of employed bees is equal to the number of food sources around the hive. The onlooker bee chooses probabilistically a food source depending on the amount of nectar shown by each employed bee, see equation (15).

$$P_i = \frac{fit_i}{\sum_{n=1}^{SN} fit_i} \quad (15)$$

Where $fit_i$ is the fitness value of the solution $i$ and $SN$ is the number of food sources which are equal to the number of employed bees.

Each bee searches for new neighbor food source near of their hive. After that, employed bee compares the food source against the old one using equation 16. Then, it saves in their memory the best food source [13-14].

$$V_{ij}^{new} = x_{ij}^{old} + \varphi_{ij}(x_{ij}^{old} - x_{kj}^{old}) \tag{16}$$

Where $k \in \{1,2,3,...,NP\}$ and $j \in \{1,2,3,...,n\}$ are chosen randomly. $\varphi_{ij}$ is a random number between [-1, 1]. In ABC, Providing that a position cannot be improved further through a predetermined number of cycles, then that food source is assumed to be abandoned. The value of predetermined number of cycles is an important control parameter of the ABC algorithm, which is called "limit" for abandonment. Assume that the abandoned source is $x_i$ and $j= \{1, 2... N\}$ then the scout discovers a new food source to be replaced with $x_i$. This operation can be defined using equation 17.

$$x_i^j = x_j^{min} + rand(0,1)(x_j^{max} - x_j^{min}) \tag{17}$$

ABC implementation stages of the algorithm:
1. Initialization
2. Place employed bees on the food sources
3. Place onlookers on the food sources
4. Send scout to carry out the global search

Flowchart of the ABC algorithm is shown in figure4 [12].

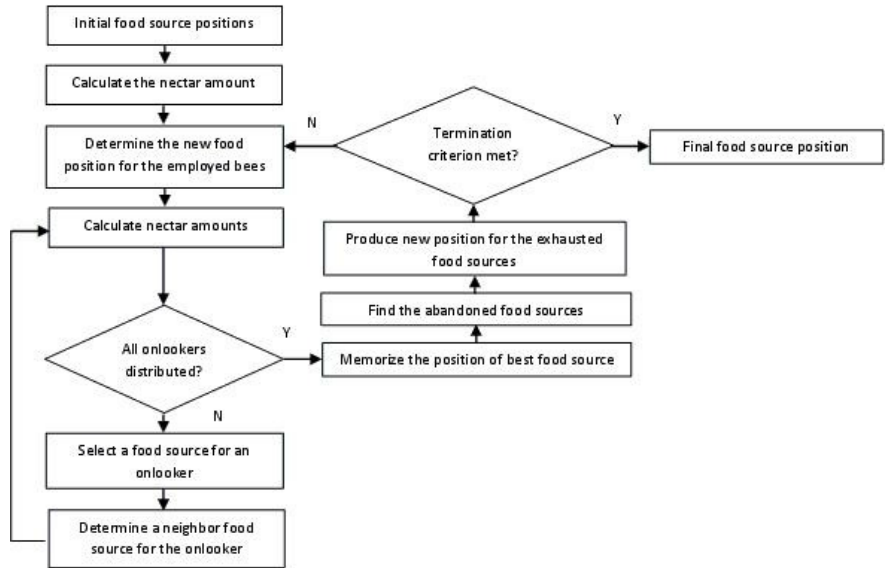

Figure 4. Flowchart of the ABC algorithm

## 3.3. Hybrid of Genetic Algorithm and Particle Swarm Optimization (HGAPSO)

Concept of GA:

Genetic algorithm was introduced by John Holland in 1967. Later, in1989, by Goldberg's attempts got accepted. And today, due to its capability, it has prosperous place among other methods. Routine optimization in genetic algorithm bases on random- directed procedure. In this way depending on Darwin's theory of evolution and fundamental ideas the method has been firmed. First of all, for some constants which are called population, a swarm of main parameters, eventually get produced. After performing scalar simulator program, a number that is attributed as standard deviation or set fitness data, impute as a member of that population. Fitness process is done for all created members then by summoning genetic algorithm operators such as succession (crossover), mutation and next generation selection (selection) are formed. This routine continues till satisfying convergence [15].

Concepts of PSO:

Initially, velocity and position of each particle randomly select and direct all particles gradually to the best answer. PSO has high speed in local and total search. Applying PSO takes to dominate rapid and untimely convergence of GA that is achieved with elitism operations. Also, PSO can fit particles of each generation whereas GA is seeking for an answer generation to generation. Depending on these points, GA and PSO are good supplements of each other [16].

The HGAPSO algorithm is following as bellow:
1. A population randomly should be selected, initially, which is parameters of GA and particles of PSO.
2. The fitness of concluding is calculating.
3. Half of population with high fitness would be selected and then PSO function would be performed. But a factor breeding ratio was presented with φsymbol [17]. The factor determines the percentage of population on which the PSO operation is done.
4. Enhanced elites would be transmitted to the next generation directly and mutation and remixing functions would be done on the rest population.

Figure 5 shows all the above steps and flowchart of the HGAPSO algorithm is shown in Figure6 [16].

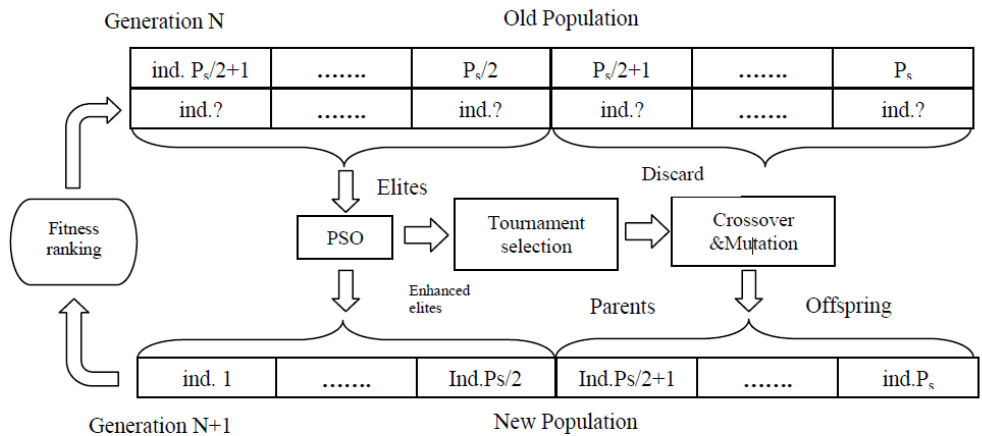

Figure5. Steps of HGAPSO algorithm [16]

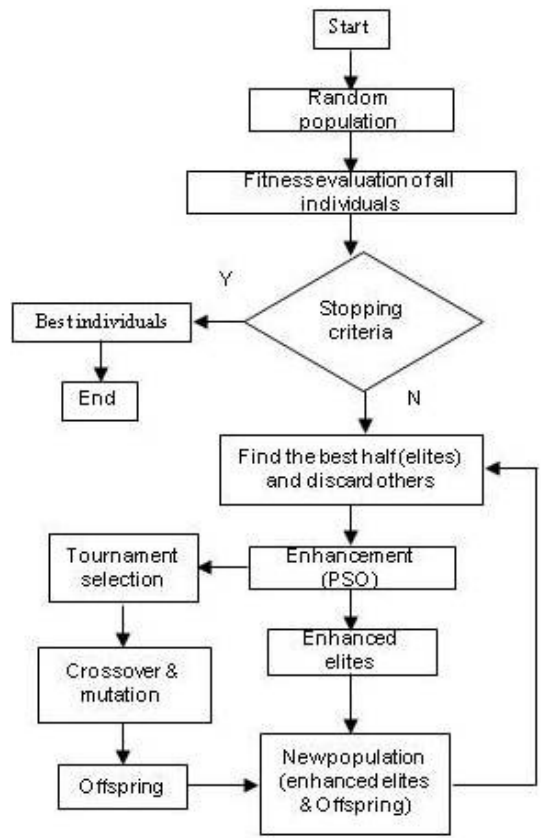

Figure 6. Flowchart of the HGAPSO algorithm

## 4. RESULTS AND DISCUSSION

In this section the results of mentioned algorithms implementation in our optimization problem for minimizing cost function will be presented and discussed. Table 1 expresses the results of dynamic balancing optimization with PSO, ABC, BGA and HGAPSO algorithms for 200 and 300 iterations. Also programming is done with MATLAB12, with 2.8 GHZ computer system processor and 1024MB memory. By these tables, best and worst average results are shown in bold. Results comparison indicates that PSO algorithm is more efficient than other algorithms, as the average cost of objective function for 20 iterations is 29853.35062, and for 300 iterations is 29853.30955 that is less than three other algorithms.

Table 1. Results of dynamic balancing optimization with PSO, ABC, BGA and HGAPSO algorithms for 200 and 300 iterations

**Statistical Results Obtained by the BGA algorithm for 200 Generation**

| Experiment | M1 | M2 | Fi1 | Fi2 | Cost | CPU Time |
|---|---|---|---|---|---|---|
| #1 | 425.2199 | 770.088 | 2.260227 | 4.907395 | 29917.7403 | 27.816430 |
| #2 | 200 | 312.61 | 3.144664 | 4.710854 | 30037.9671 | 29.935968 |
| #3 | 425.2199 | 648.6804 | 2.395349 | 4.944247 | 29982.0926 | 28.821098 |
| #4 | 305.5718 | 374.1935 | 3.722004 | 5.220633 | 29854.0986 | 28.366290 |
| #5 | 312.61 | 648.6804 | 2.352356 | 4.876685 | 29914.4407 | 32.985386 |
| #6 | 368.915 | 789.4428 | 1.228384 | 4.391474 | 29928.107 | 32.076005 |
| #7 | 1183.5777 | 1099.1202 | 2.9419802 | 5.7979735 | 30043.6419 | 28.184769 |
| #8 | 555.4252 | 310.8504 | 3.924688 | 6.04365 | 29876.8727 | 28.312547 |
| #9 | 425.2199 | 873.9003 | 1.959273 | 4.802982 | 29867.9256 | 31.578058 |
| #10 | 536.0704 | 407.6246 | 5.312762 | 3.138522 | 29895.7188 | 30.211240 |
| Average | | | | | 29931.86 | 29.82878 |
| Best | | | | | 29854.1 | 27.81643 |
| Worst | | | | | 30043.64 | 32.98539 |

**Statistical Results Obtained by the ABC algorithm for 200 Iterations**

| Experiment | M1 | M2 | Fi1 | Fi2 | Cost | CPU Time |
|---|---|---|---|---|---|---|
| #1 | 957.917 | 878.0169 | 9.631981 | 4.870459 | 29853.8571 | 24.515309 |
| #2 | 458.3323 | 857.7992 | 8.598309 | 5.188452 | 29854.7879 | 24.497529 |
| #3 | 1131.1942 | 1101.4885 | 9.2855979 | 1.8518603 | 29856.3607 | 25.638495 |
| #4 | 217.0502 | 221.4425 | 10.7222 | 13.26839 | 29854.8753 | 25.814662 |
| #5 | 260.4993 | 623.3153 | 8.486066 | 11.19885 | 29855.1337 | 26.448376 |
| #6 | 677.15 | 321.5682 | 10.37823 | 6.049843 | 29854.5122 | 28.761479 |
| #7 | 1237.6854 | 960.86413 | 10.019323 | 5.8620878 | 29856.5621 | 27.566192 |
| #8 | 746.1877 | 749.8799 | 6.339937 | 9.548391 | 29854.274 | 27.581421 |
| #9 | 639.2974 | 315.6203 | 7.741611 | 12.66076 | 29855.4856 | 27.869337 |
| #10 | 683.5569 | 888.9545 | 2.799224 | 3.648658 | 29854.6395 | 27.622605 |
| Average | | | | | 29855.04881 | 26.6315405 |
| Best | | | | | 29853.8571 | 24.497529 |
| Worst | | | | | 29856.5621 | 28.761479 |

**Statistical Results Obtained by the BGA algorithm for 300 Generation**

| Experiment | M1 | M2 | Fi1 | Fi2 | Cost | CPU Time |
|---|---|---|---|---|---|---|
| #1 | 585.3372 | 416.4223 | 3.777281 | 5.94538 | 29860.512 | 69.617549 |
| #2 | 537.8299 | 845.7478 | 0.393083 | 3.881694 | 29979.2346 | 75.797888 |
| #3 | 363.6364 | 536.0704 | 3.040251 | 5.208349 | 29900.57 | 70.299023 |
| #4 | 620.5279 | 0.7861659 | 4.121229 | 256.305 | 29988.9287 | 61.185702 |
| #5 | 397.0674 | 648.6804 | 2.745439 | 5.128504 | 29896.1839 | 67.436636 |
| #6 | 321.4076 | 529.0323 | 5.920812 | 3.90012 | 29866.535 | 73.210396 |
| #7 | 228.1525 | 690.9091 | 1.940847 | 4.710854 | 29866.4603 | 85.515155 |
| #8 | 200 | 564.2229 | 1.670603 | 4.465177 | 29976.0744 | 67.414340 |
| #9 | 268.6217 | 310.8504 | 3.924688 | 5.067085 | 29864.7926 | 66.666551 |
| #10 | 485.044 | 423.4604 | 3.531605 | 5.650567 | 29900.5253 | 77.981415 |
| Average | | | | | 29909.98 | 71.41247 |
| Best | | | | | 29860.51 | 61.1857 |
| Worst | | | | | 29988.93 | 85.51516 |

**Statistical Results Obtained by the ABC algorithm for 300 Iterations**

| Experiment | M1 | M2 | Fi1 | Fi2 | Cost | CPU Time |
|---|---|---|---|---|---|---|
| #1 | 889.97288 | 1026.2293 | 6.8552951 | 13.822746 | 29856.0266 | 40.422035 |
| #2 | 374.24233 | 1405.1666 | 7.839461 | 10.87474 | 29859.7044 | 42.796960 |
| #3 | 260.895 | 501.8795 | 7.752119 | 10.27209 | 29854.4018 | 40.764053 |
| #4 | 1624.3941 | 1153.022 | 10.633852 | 7.5664349 | 29855.7171 | 43.498866 |
| #5 | 843.7006 | 359.6281 | 11.00391 | 8.106481 | 29854.1593 | 45.076684 |
| #6 | 772.2693 | 653.4779 | 5.531582 | 9.657654 | 29854.0499 | 48.079070 |
| #7 | 2632.9495 | 1494.202 | 4.5137612 | 7.6742805 | 29854.0659 | 48.203408 |
| #8 | 1440.1067 | 1182.0524 | 10.868796 | 9.1406666 | 29854.8792 | 47.887119 |
| #9 | 782.1371 | 569.479 | 9.691396 | 6.195521 | 29854.4519 | 49.274214 |
| #10 | 655.3956 | 612.1464 | 9.454865 | 5.618574 | 29853.7587 | 45.614997 |
| Average | | | | | 29855.22148 | 45.1617406 |
| Best | | | | | 29853.7587 | 40.422035 |
| Worst | | | | | 29859.7044 | 49.274214 |

**Statistical Results Obtained by the PSO algorithm for 200 Iterations**

| Experiment | M1 | M2 | Fi1 | Fi2 | Cost | CPU Time |
|---|---|---|---|---|---|---|
| #1 | 945.7894 | 457.6034 | 4.413292 | 7.388778 | 29853.3719 | 13.877018 |
| #2 | 464.4661 | 715.8224 | 9.17736 | 5.266925 | 29853.3846 | 13.174372 |
| #3 | 83.58313 | 483.6949 | 5.854242 | 10.68152 | 29853.3894 | 12.264391 |
| #4 | 142.892 | 479.2902 | 6.42547 | 11.13792 | 29853.4599 | 12.552338 |
| #5 | 190.3871 | 344.1711 | 7.216235 | 11.11191 | 29853.3813 | 12.357725 |
| #6 | 112.7041 | 583.2056 | 6.888962 | 10.70696 | 29853.305 | 13.300798 |
| #7 | 156.9876 | 467.6164 | 5.76949 | 10.52967 | 29853.2973 | 13.855493 |
| #8 | 1201.8414 | 738.00028 | 4.2686516 | 10.352361 | 29853.2971 | 14.755708 |
| #9 | 527.3684 | 161.494 | 10.53725 | 6.146666 | 29853.3101 | 13.943058 |
| #10 | 766.8097 | 318.9964 | 11.13244 | 8.445869 | 29853.3096 | 12.585938 |
| Average | | | | | 29853.35062 | 13.2666839 |
| Best | | | | | 29853.2971 | 12.264391 |
| Worst | | | | | 29853.4599 | 14.755708 |

**Statistical Results Obtained by the HGAPSO algorithm for 200 Generation**

| Experiment | M1 | M2 | Fi1 | Fi2 | Cost | CPU Time |
|---|---|---|---|---|---|---|
| #1 | 1319.7893 | 1775.2657 | 6.7011554 | 6.7630704 | 30080.102 | 32.479893 |
| #2 | 352.3594 | 690.4537 | 9.285448 | 9.868578 | 29960.9037 | 34.840757 |
| #3 | 476.5722 | 573.7634 | 9.527518 | 9.563268 | 30036.1098 | 34.614229 |
| #4 | 828.16894 | 1221.9607 | 8.309838 | 9.2001952 | 29943.6233 | 34.057494 |
| #5 | 1841.2912 | 1609.2847 | 6.4433827 | 9.4164658 | 29994.6861 | 33.216922 |
| #6 | 1508.8175 | 772.7226 | 7.3714832 | 7.1583033 | 30080.1051 | 33.737740 |
| #7 | 1036.9438 | 1116.5328 | 7.1023018 | 9.321701 | 30033.1305 | 35.140728 |
| #8 | 1494.2831 | 632.94301 | 9.5999394 | 9.6251599 | 30069.7696 | 34.970999 |
| #9 | 763.2089 | 318.8483 | 7.74299 | 7.563557 | 30040.2645 | 34.003678 |
| #10 | 1129.1558 | 619.2222 | 9.6453097 | 6.6337606 | 30060.2337 | 35.567686 |
| Average | | | | | 30029.89283 | 34.2630126 |
| Best | | | | | 29943.6233 | 32.479893 |
| Worst | | | | | 30080.1051 | 35.567686 |

**Statistical Results Obtained by the PSO algorithm for 300 Iterations**

| Experiment | M1 | M2 | Fi1 | Fi2 | Cost | CPU Time |
|---|---|---|---|---|---|---|
| #1 | 393.2882 | 107.7893 | 7.67673 | 14.09963 | 29853.3602 | 21.686771 |
| #2 | 244.4886 | 443.838 | 9.758278 | 8.21717 | 29853.297 | 21.145141 |
| #3 | 388.6466 | 160.5542 | 11.11079 | 7.028177 | 29853.2968 | 20.927836 |
| #4 | 384.6346 | 115.9811 | 10.82956 | 7.769524 | 29853.3014 | 20.836519 |
| #5 | 262.538 | 702.0516 | 5.413787 | 7.99123 | 29853.3227 | 21.101022 |
| #6 | 360.6737 | 355.9938 | 8.498214 | 6.902059 | 29853.3127 | 20.567431 |
| #7 | 428.9131 | 143.0469 | 7.97523 | 6.787767 | 29853.2976 | 22.235461 |
| #8 | 349.7185 | 152.8699 | 10.91113 | 10.70469 | 29853.2995 | 21.991660 |
| #9 | 1370.8073 | 960.25814 | 10.353903 | 6.9624157 | 29853.2968 | 22.241028 |
| #10 | 493.1262 | 83.99186 | 83.99186 | 5.966398 | 29853.3108 | 21.751784 |
| Average | | | | | 29853.30955 | 21.4484653 |
| Best | | | | | 29853.2968 | 20.567431 |
| Worst | | | | | 29853.3602 | 22.241028 |

**Statistical Results Obtained by the HGAPSO algorithm for 300 Generation**

| Experiment | M1 | M2 | Fi1 | Fi2 | Cost | CPU Time |
|---|---|---|---|---|---|---|
| #1 | 772.9007 | 595.1176 | 6.86203 | 7.937233 | 30073.0299 | 51.955371 |
| #2 | 882.80336 | 1896.9992 | 7.2606345 | 9.959396 | 29955.7353 | 55.753747 |
| #3 | 375.9766 | 405.99 | 7.360915 | 9.92088 | 30037.4567 | 55.435066 |
| #4 | 355.42156 | 1227.6115 | 9.7068992 | 9.2411975 | 29982.7794 | 55.902381 |
| #5 | 1572.9019 | 874.43037 | 7.0439925 | 8.9360977 | 30068.9326 | 55.842394 |
| #6 | 398.9521 | 827.0116 | 7.097475 | 8.669514 | 30007.8626 | 54.642494 |
| #7 | 1411.5414 | 574.13944 | 8.6237598 | 6.9623394 | 29936.2702 | 56.950295 |
| #8 | 1856.2111 | 871.64681 | 7.7394652 | 9.7647663 | 30046.7941 | 56.097725 |
| #9 | 1827.0208 | 321.74463 | 8.1571052 | 9.2847123 | 29968.9997 | 56.730746 |
| #10 | 770.12423 | 1724.5737 | 9.4666394 | 8.4071409 | 29951.5584 | 55.731429 |
| Average | | | | | 30002.94189 | 55.5041648 |
| Best | | | | | 29936.2702 | 51.955371 |
| Worst | | | | | 30073.0299 | 56.950295 |

Also, above mentioned results in the table are illustrated in Figures 7 and 8 in bar chart which show the comparison of objective function cost. By observing these figures, it may be concluded that the cost order is as following: PSO<ABC<BGA<HGAPSO.

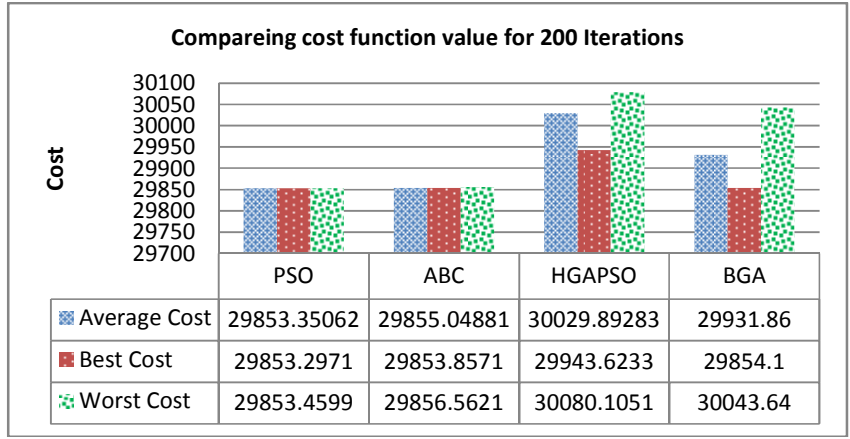

Figure7. Comparing cost function value for 200 Iterations

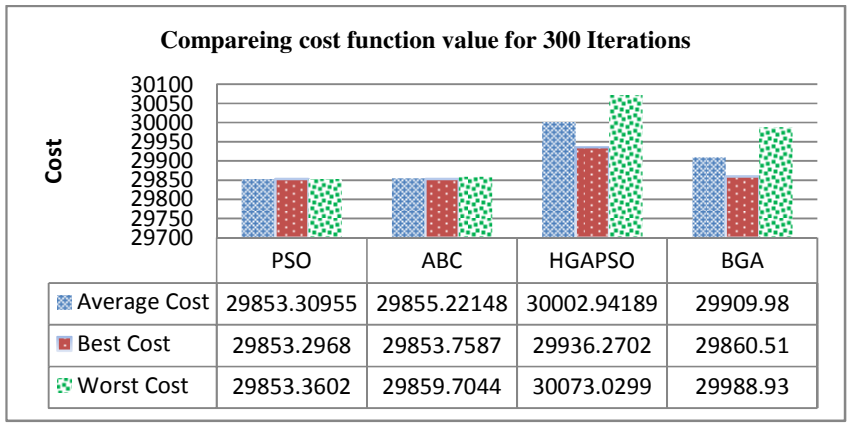

Figure8 . Comparing cost function value for 300 Iterations

In addition, the runtime comparison shows that PSO algorithm finds the optimal solution in shorter time than the others. Also the average runtime for this algorithm for 200 iterations is 13.2666839 seconds and for 300 iterations is 21.4484653 seconds. According to numerical results PSO algorithm is 3.84 times faster than BGA and 2.56 times faster than HGAPSO and 2.21 times faster than ABC. Figures 9 and 10 indicate the runtime of these algorithms for one performance with 200 and 300 iterations, respectively. By observing these figures, it maybe found that the runtime order is as following: PSO<ABC<HGAPSO<BGA. It should be mentioned that the runtime for Hybrid algorithm of BGA and PSO algorithm is 1.5 times faster than BGA algorithm in reaching optimal solution.

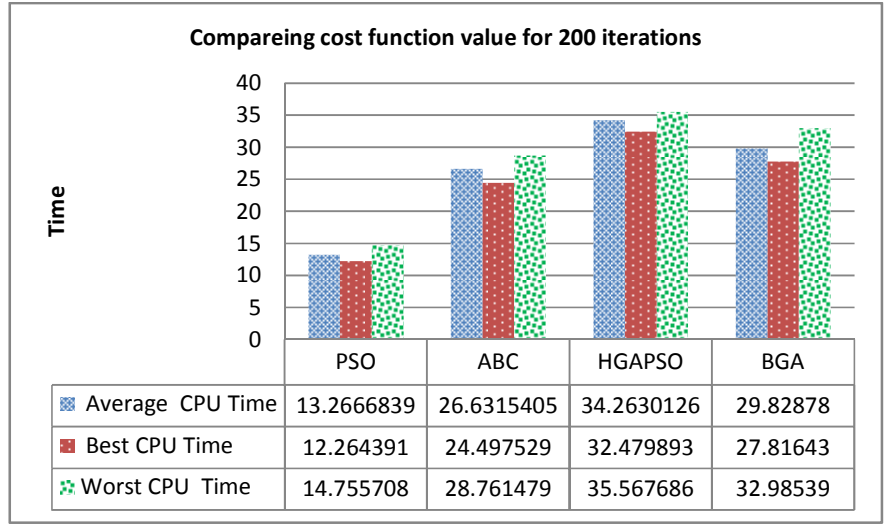

Figure 9. Comparing runtime for 200 iterations

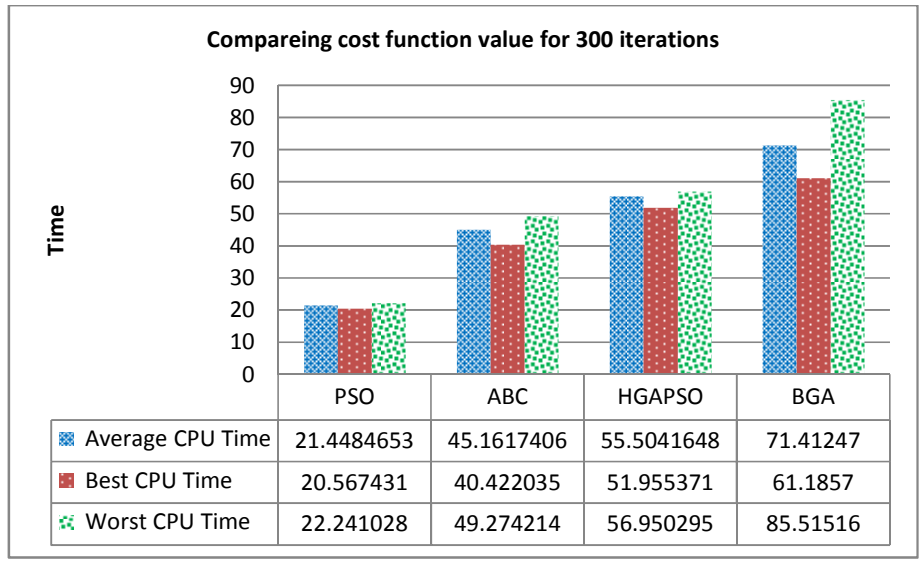

Figure 10. Comparing runtime for 300 iterations

Figure 11 indicates the time order of these algorithms for one performance with 100 and 200 repetitions. In this chart, the implementation of time progress of these algorithms is in term of seconds.

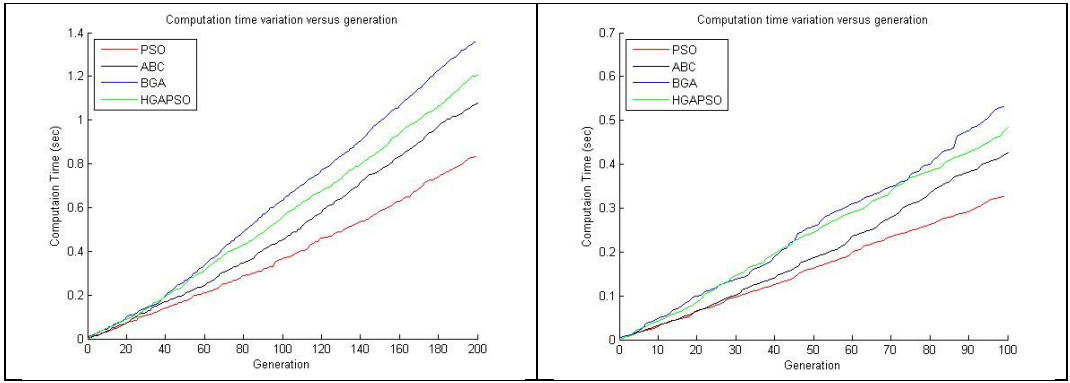

Figure 11. Computation time variation versus generation for 100 iterations
(Left) and 200 iterations (right)

Figure 12 shows fitness functions of four algorithms in 100 runs. In this figure the range of fluctuation of fitness value for BGA and HGAPSO algorithms is between 29853 and 29855, while this range for BGA and HGAPSO is approximately between 29853 and 30200. Also high difference between the best and worst result of the BGA and HGAPSO algorithms indicates the weakness and dependence of these two algorithms and their sensitivity to the initial population, while PSO and ABC solution range is around the mean values, so it indicates the stability of them. Also, the PSO and ABC algorithms are close to average result that this matter indicates the stability of these two algorithms.

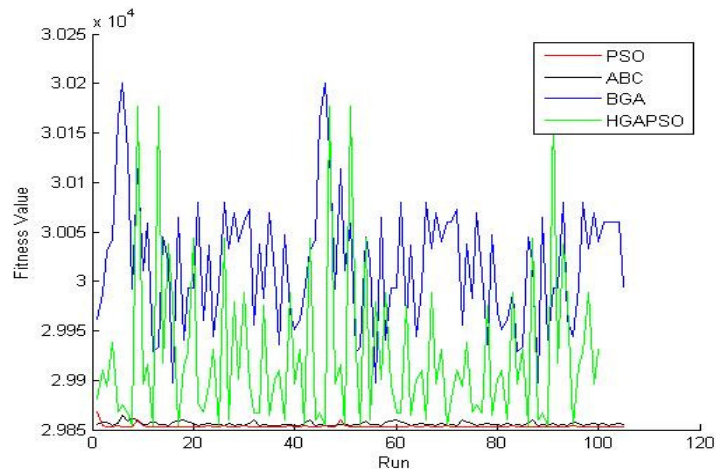

Figure12. Fitness functions fluctuation for 100 runs for four algorithms

Figure 13 and 14 display the four algorithms convergence in a sample run of them (for 200 and 300 iterations). According to the minimization target, an important point which should be considered is that the charts are descending. In addition the scale of $10^4$ is using for measuring of objective function. As comparing of these two figures, it may be observed that the PSO and ABC algorithms get convergence approximately after 30 generations also they achieve the optimized solution in high speed rather than HGAPSO and GA.

International Journal of Artificial Intelligence & Applications (IJAIA), Vol. 4, No. 5, September 2013

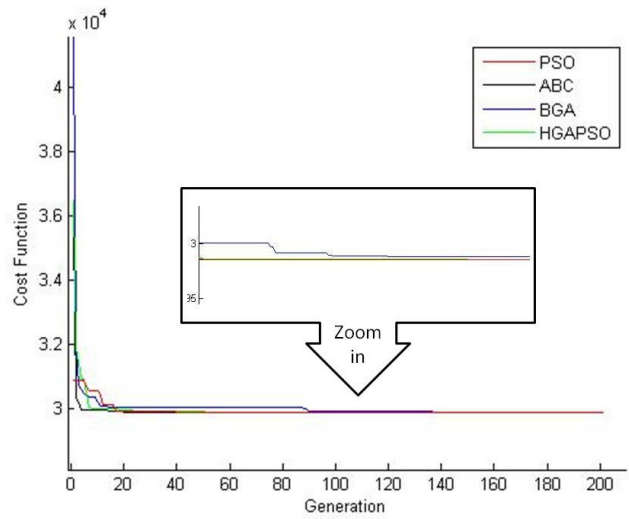

Figure13.Convergence plot for 200 iterations

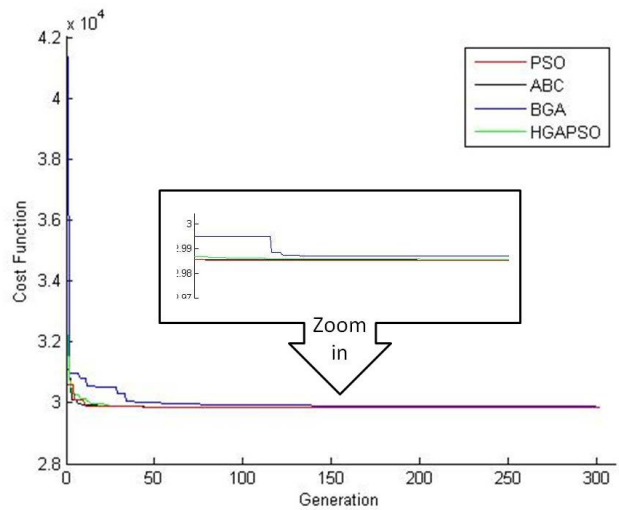

Figure14.Convergence plot for 300 iterations

Finally, Figure15 indicates cost function diagram in balanced and unbalanced states by mentioned algorithms in polar plot. This type of plot is a standard plot which applied for illustrating the ability of the proposed method in balancing of the mechanism in one revolution of the mechanism driver (0-360°). As it was described in previous sections, the main purpose of optimization is minimizing of the cost function in equation 9 which maybe interpreted as minimizing the limited area in the mentioned plot. Therefore better balancing method is a method which has minimum and uniform area in the polar plot. By observing this figure, it maybe concluded that PSO has better performance than other algorithm, because it contains minimum area in comparing with other algorithms.

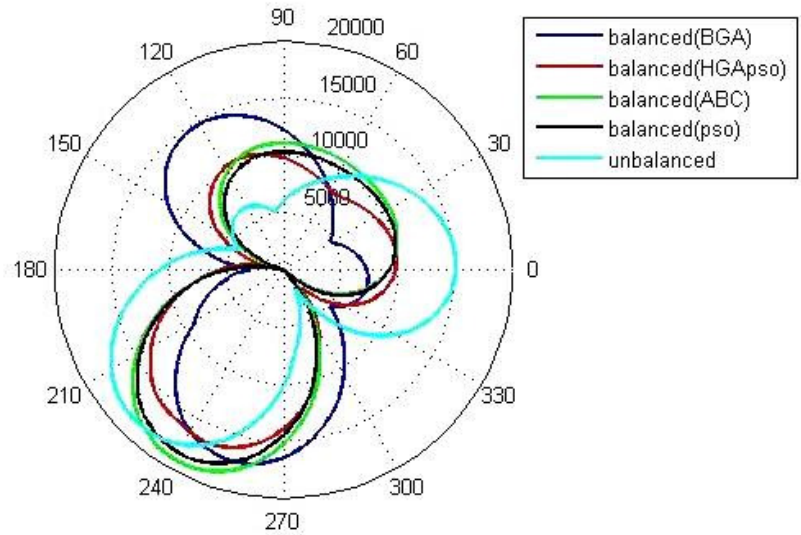

Figure 15. Polar plot of cost function for unbalanced and balanced states by applying different algorithms

## 5. EXPERIMENTALLY VALIDATION OF THE PROPOSED METHOD

For showing the practicability of the proposed balancing method based on mentioned optimization algorithms, a laboratory scale experimental double four-bar crank-slider mechanism was provided as shown in Figure16 [10].By using two accelerometers (type 4507 B&K Co.) and data acquisition system (PULSE B&K Co.), the acceleration of the right and left bearing of the mechanism can be extracted. Also a tachometer (Type 0024 MM B&K Co.) is applied for identifying the one revolution of the crank for estimating the angular acceleration. As a sample, the acceleration signals of the unbalanced and balanced mechanism, applying BGA, are illustrated in Figures 17 and 18. By comparing the mentioned linear and angular acceleration signal of the unbalanced and balanced mechanism, the possibility and accuracy of the proposed method in balancing of real mechanisms can be observed. In addition the corresponding angular acceleration signals are illustrated in Figure 19. As it is obvious from these figures, the proposed method is able to balance the mechanism because the overall amplitude of the linear and angular acceleration was decreased after balancing. Other similar results may be obtained by applying other algorithms, which cannot be illustrated in this paper because of the page limitation. By considering the experimental result, it can be observed that the same conclusion maybe derived as in previous section while comparing these algorithms.

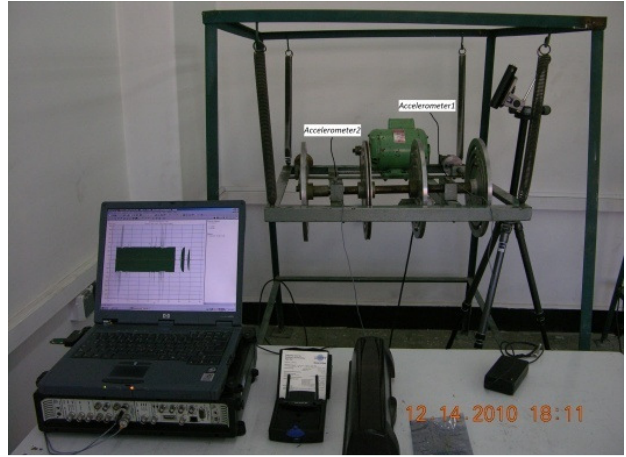

Figure16.A laboratory scale experimental double four-bar crank-slider mechanism setup

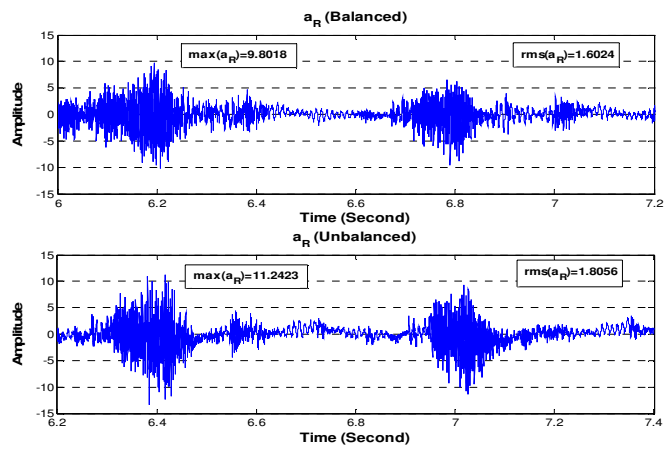

Figure17. Acceleration signal of the right part of the mechanism for before and after balancing

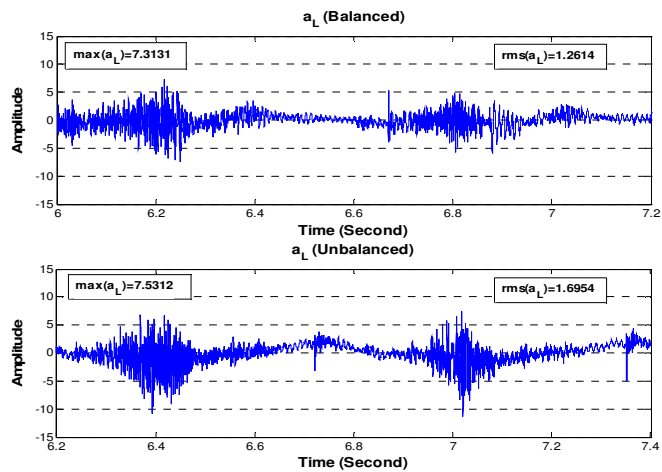

Figure18.Acceleration signal of the left part of the mechanism for before and after balancing

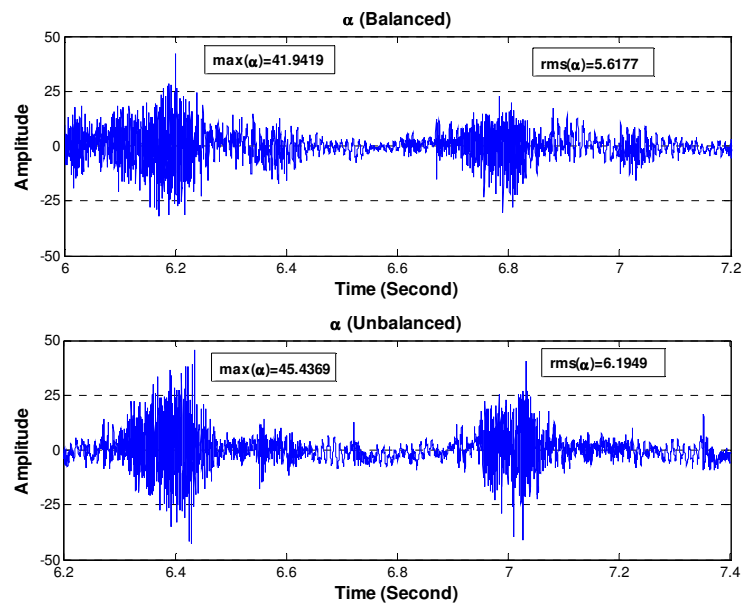

Figure 19. Angular acceleration signal of the mechanism for before and after balancing

## 6. CONCLUSION

In this paper, four Meta-Heuristic algorithms, that are PSO, ABC, BGA and HGAPSO, were applied for balancing of double four-bar crank slider mechanism. This mechanism is a basic and benchmark mechanism which is applied in different machine and engines. Therefore proposed balancing method may be applied for different type of machines for reducing the vibration and noise because of unbalancing. Kinematic modelling of the mechanism and suitable objective function which should be minimized and the constraints are derived. In order to comparing the optimization algorithm, different results based on cost function, fitness, convergence speed and runtime values were extracted and studied. The results show that the PSO algorithm converges quickly to the optimum solution, while the optimality of PSO solutions has a higher quality than other algorithms. Also, reviewing the responses of the four algorithms in several performances with various primary populations determined that, PSO and ABC are more stable than BGA and HGAPSO. Therefore PSO and ABC have a good potential in finding optimal solution and its convergence characteristics are more favourable. In addition, the practicability of the method was proved by applying the proposed method on a laboratory scale experimental complex crank slider mechanism.

## REFERENCES

[1] A.A. Sherwood & B.A. Hockey, (1968)"The optimization of mass distribution in mechanisms using dynamically similar systems", Journal of Mechanisms, Vol. 4, pp. 243–260.
[2] G.G. Lowen & R.S. Berkof, (1971) "Determination of forced-balance four-bar linkages with optimum shaking moment characteristics", ASME Journal of Engineering for Industry, Vol. 93, No. 1, pp. 39–46.
[3] H. Chaudhary & S. K. Saha, (2007) "Balancing of four-bar linkages using maximum Recursive dynamic algorithm", Mechanism and Machine Theory, Vol. 42, pp. 216–232.
[4] N. M. QI & E. Pennestrl, (1991) "Optimum balancing of four-bar linkage, a Refined algorithm," Mech. Mach. Theory, Vol. 26, No. 3, pp. 337–348.

[5] Y. YU, (1987) "Optimum shaking force and shaking moment balancing of the RSS'R spatial linkage," Mech. Mac& Theory, Vol. 22, No. I, pp. 39–45.

[6] G. Alici & B. Shirinzadeh, (2006) "Optimum dynamic balancing of planar parallel manipulators based on sensitivity analysis," Mechanism and Machine Theory, Vol. 41, pp. 1520–1532.

[7] Wen-Yi Lin, (2010) " A GA–DE hybrid evolutionary algorithm for path synthesis of four-bar linkage, " Mechanism and Machine Theory, Vol.45, No.8, pp. 1096-1107

[8] S. K. Acharyya & M. Mandal, (2009)"Performance of EAs for four-bar linkage synthesis ", Mechanism and Machine Theory, Vol.44, pp.1784-1794.

[9] A. Jaamiolahmadi & M. R. Farmani, (2006) "Optimal force and moment balance of a four bar linkage via genetic algorithm," 14th Annual (International) Mechanical Engineering Conference-Esfahan - Iran.

[10] M. M. Ettefagh, F.abbasidust, H.milanchian & M.yazdanian (2011) "Complex Crank-Slider Mechanism Dynamic Balancing by Binary Genetic Algorithm (BGA)", symposium in Innovations in Intelligent Systems and Applications (INISTA2011) - Istanbul, pp.277 – 281.

[11] J. Kennedy & R. Eberhart, (1995). "Particle Swarm Optimization," Proceeding of IEEE International Conference on Neural Networks in Perth-WA, Vol.4, pp. 1942-1948.

[12] D. Karaboga & B. Basturk, (2008) "On the performance of artificial bee colony (ABC) algorithm," Applied Soft Computing, Vol. 8, pp. 687–697.

[13] S.H. Sishaj & P. Simon, (2010) "Artificial bee colony algorithm for economic load dispatch problem with non-smooth cost functions," Electric Power Components and Systems, Vol. 38, No. 7, pp. 786-803.

[14] F. Kang, J. Li , H. Li, Z. Ma & Q. Xu, (2010) "An improved artificial bee colony algorithm", Intelligent Systems and Applications (ISA), pp.1-4.

[15] Z. Michalewicz, (1999) "Genetic Algorithms+ Data Structures=Evolution Programs", Springer-Verlag Berlin Heidelberg New York.

[16] Chia-Feng Juang, (2004) "A Hybrid of Genetic Algorithm and Particle swarm Optimization for Recurrent Network Design", IEEE Transactions on system, Man and Cybernetics, Vol. 34,No. 2, pp.997-1006.

[17] M.Settles, P. athan & T. Soule, (2005)" Breeding Swarms: A New Approach to Recurrent Neural Network Training", Proceedings of the 2005 conference on Genetic and evolutionary computation (GECCO 2005), pp. 185-192.

**Authors**

Mir Mohammad Ettefagh received the PhD degree in Mechanical Engineering from the University of Tabriz in 2007. He is currently an Assistant professor in Mechanical Engineering department of the University of Tabriz. His research interests include vibration and dynamic systems, optimization of mechanism, fault diagnosis of machines, structural health monitoring.

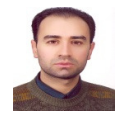

Mohammad-Reza Feizi-Derakhshi, Ph.D. is currently a faculty member at the University of Tabriz. His research interests include: natural language processing, optimization algorithms, intelligent methods for fault detection and intelligent databases.

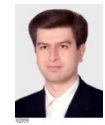

Habib Emdadi received his BS degree in Software Engineering Technology from University of Tabriz, Iran in 2009. He is currently working toward the MS degree of Computer Science with Intelligent Systems in Tabriz University, focusing on Optimization Techniques. His research interests include Evolutionary Computation, Optimization Techniques, soft computing and Computer Programming.

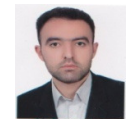

Mahsa Yazdanian received the BS degree in Mechanical Engineering from University of Tabriz, Iran in 2010. She is currently working toward the MS Degree in the Department of Mechanical Engineering. Her research interests include artificial-based optimization of mechanism and condition monitoring of rotating machinery.

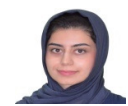